\documentclass[runningheads]{llncs}
\usepackage[T1]{fontenc}

\usepackage{amsmath,amssymb}
\usepackage{booktabs}
\usepackage{multirow}
\usepackage{graphicx,verbatim}

\begin{document}
\title{Do Foundation Models See Biology? Evaluating Attention Coherence with Spatial Transcriptomics in Glioblastoma}
\titlerunning{Do Foundation Models See Biology?}
\author{Dilakshan Srikanthan\inst{1,2}\and
Amoon Jamzad\inst{2} \and
Paul Wilson\inst{2}\and
Nooshin Maghsoodi\inst{2} \and
Robert Policelli\inst{1}\and
Gabor Fichtinger\inst{2}\and
John F. Rudan\inst{3}\and
Parvin Mousavi\inst{1,2}}

\authorrunning{D. Srikanthan et al.}

\institute{Translational Medicine, School of Medicine, Queen's University, Kingston, ON, Canada \and
School of Computing, Queen's University, Kingston, ON, Canada \and
Department of Surgery, Queen's University, Kingston, ON, Canada\\
\email{dilakshan.srikanthan@queensu.ca}}

\maketitle            % typeset the header of the contribution

\begin{abstract}
Whether attention maps from pathology foundation models capture genuine biology remains unknown, yet this question is critical for clinical trust and regulatory approval. We propose a spatial transcriptomics-based framework for orthogonal, hypothesis-free evaluation of attention and apply it to five pathology foundation models (CONCH v1.5, UNI v2, Virchow2, GigaPath, H-Optimus-1) and a ResNet50 baseline. Using attention-based multiple instance learning, we train single-task and multi-task models to predict five molecular alterations in glioblastoma on the CPTAC cohort, validate on an independent TCGA cohort, and evaluate biological coherence of attention maps against 87 transcriptional signatures using co-registered Visium spatial transcriptomics data from 18 samples. Internally, no single encoder dominates across all tasks, and external validation inverts internal performance rankings. Attention maps show a five-fold enrichment gradient from pathways (Cohen's d=0.329) to individual genes (d=0.055), indicating that attention captures emergent multi-gene transcriptional programs rather than individual molecular events. Spatially smooth attention maps do not imply biological coherence, and different encoders attend to distinct biological compartments. Our framework provides objective, quantitative assessment of what foundation models learn from histopathology, moving the field beyond qualitative saliency map review.

\end{abstract}

\section{Introduction}

Pathology foundation models have been trained on millions of whole slide images (WSI) and can now predict molecular biomarkers, tumour grade, and stratify patient outcomes across various cancer types~\cite{ref_uni,ref_virchow2,ref_conch,ref_gigapath,ref_hoptimus}. Yet as these models approach clinical deployment, a gap persists between predictive accuracy and interpretability. For clinical deployment,  transparency and explainability become essential to support trust, error detection, and regulatory approval. In digital pathology, attention-based multiple instance learning (abMIL)~\cite{ref_ilse} partially addresses this by assigning spatial weights to tissue patches, producing maps of which regions contribute most to slide-level predictions. However, whether these attention weights constitute a faithful explanation remains debated~\cite{ref_jain,ref_wiegreffe}. Alternative attribution methods have their own limitations~\cite{ref_grad}.

In practice, biological interpretation of attention maps is almost universally performed through qualitative pathologist review~\cite{ref_uni,ref_virchow2}, where experts visually assess whether highlighted regions correspond to known histological features. While valuable, this approach is subjective, prone to inter-rater variability, and cannot scale to large numbers of tissue regions that modern foundation model benchmarks require. The question we address is whether there exists a way to objectively and quantitatively assess what attention maps capture, independent of human interpretation and independent of the model itself. Spatial transcriptomics offers a possible solution by measuring genome-wide gene expression at known tissue coordinates. It provides a molecular readout that is orthogonal to the image features the model processes. If high-attention regions consistently co-localize with defined transcriptional programs measured by an entirely separate assay, the correspondence is grounded in biology rather than in circular reasoning from the model's own learned representations. To our knowledge, no study has used spatial transcriptomics in this systematic, untargeted manner to evaluate foundation model attention or any deep learning model attention ~\cite{ref_ng}.

We apply this framework to glioblastoma (GBM), where molecular heterogeneity drives classification and treatment under the 2021 WHO CNS guidelines~\cite{ref_who}, benchmarking five foundation models plus a ResNet50 baseline across five molecular targets with external validation. Our contributions are threefold.
%rephrase the independent method sentence. 
\textbf{(1)}  We provide a hypothesis-free attention evaluation framework to reveal at what level of biological organization attention operates, whether it is individual genes, cell-type markers, or coordinated multi-gene programs.
\textbf{(2)} Evidence that models that produce spatially smooth, visually coherent attention maps do not necessarily attend to biologically meaningful regions, and our benchmarking context indicates encoder performance rankings based on internal cross-validation do not hold true in an external cohort, demonstrating that neither visual inspection nor single-cohort benchmarking adequately evaluates explainability.
\textbf{(3)} Demonstration that the proposed framework characterizes what attention encodes and how it differs across deep learning models, providing a principled, quantitative basis for understanding and comparing model representations.
\section{Material and Methods}
\begin{figure}[!t]
\centering
\includegraphics[width=\textwidth]{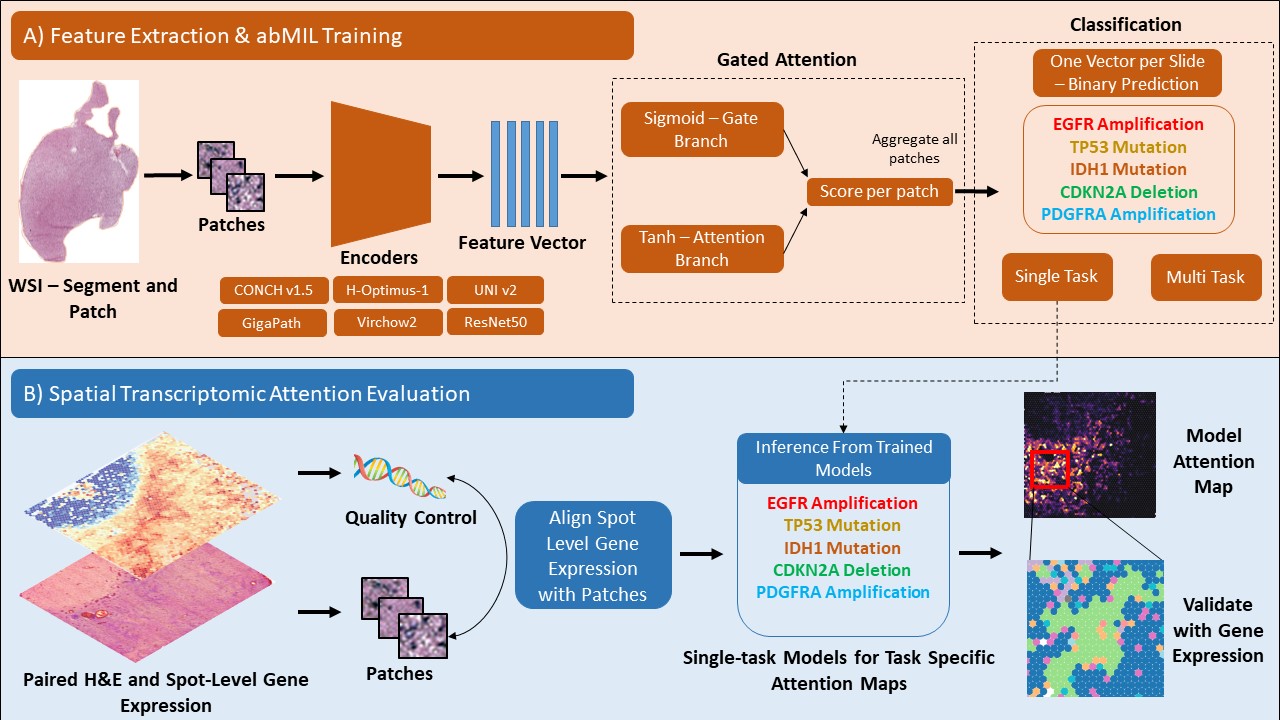}
\caption{Overview. \textbf{(A)}~Six frozen encoders produce attention maps via gated abMIL for five molecular targets. \textbf{(B)}~Co-registered Visium data quantifies biological enrichment in high- vs.\ low-attention regions.}
\end{figure}
\subsection{Cohorts and Preprocessing}
\noindent\textbf{Training cohort.}
H\&E WSIs from the CPTAC-GBM cohort ($N{=}171$ patients, 433 slides) were obtained from The Cancer Imaging Archive (TCIA) and cBioPortal~\cite{ref_cbio}. Five molecular targets were selected for single-task prediction: EGFR amplification, TP53 mutation, IDH1 mutation, CDKN2A homozygous deletion, and PDGFRA amplification, with four additional targets (PDGFRA mutation, PIK3CA mutation, PTEN mutation, NF1 mutation) for multi-task. Copy-number alterations were defined via GISTIC discrete calls (amplification: $+2$; deep deletion: $-2$)~\cite{ref_gistic}.

\noindent\textbf{External validation cohort.}
TCGA-GBM ($N{=}107$ patients, 312 slides) served as the independent test set. Trained models were applied without retraining. Patient-level AUROC with bootstrap 95\% CIs were computed.

\noindent\textbf{Preprocessing.}
Tissue was segmented with DeepLabV3~\cite{ref_deeplabv3} and divided into non-overlapping $224{\times}224$ patches at $20{\times}$ magnification ($0.5\;\mu\text{m/pixel}$), stored in H5 format.
\noindent\textbf{Feature Extraction.}
Each patch $x_i$ was encoded by a frozen feature extractor $f_\theta$ into an embedding $\mathbf{h}_i = f_\theta(x_i) \in \mathbb{R}^d$, where $d$ varies by encoder. Six encoders were evaluated: CONCH~v1.5 ($d{=}768$), UNI~v2 ($d{=}1536$), Virchow2 ($d{=}2560$), GigaPath ($d{=}1536$), H-Optimus-1 ($d{=}1536$), and a ResNet50 ImageNet baseline ($d{=}1024$). All encoders processed identical patches to ensure fair comparison.
\subsection{Attention-Based Multiple Instance Learning}
Given a WSI represented as a bag of $N$ patch embeddings $\mathcal{B} = \{\mathbf{h}_1, \ldots, \mathbf{h}_N\}$, we first project each embedding into a shared latent space:
\begin{equation}
\mathbf{z}_i = \text{ReLU}\bigl(\text{LN}(\mathbf{W}_{\text{proj}}\,\mathbf{h}_i + \mathbf{b}_{\text{proj}})\bigr) \in \mathbb{R}^{256},
\label{eq:projection}
\end{equation}
where $\text{LN}(\cdot)$ denotes LayerNorm and dropout ($p{=}0.4$) is applied. Attention weights are computed via a gated mechanism~\cite{ref_ilse}:
\begin{equation}
a_i = \frac{\exp\bigl\{\mathbf{w}^\top \bigl(\tanh(\mathbf{V}\mathbf{z}_i) \odot \sigma(\mathbf{U}\mathbf{z}_i)\bigr)\bigr\}}{\sum_{j=1}^{N} \exp\bigl\{\mathbf{w}^\top \bigl(\tanh(\mathbf{V}\mathbf{z}_j) \odot \sigma(\mathbf{U}\mathbf{z}_j)\bigr)\bigr\}},
\label{eq:attention}
\end{equation}
where $\mathbf{V}, \mathbf{U} \in \mathbb{R}^{128 \times 256}$, $\mathbf{w} \in \mathbb{R}^{128}$, $\odot$ denotes element-wise multiplication, and $\sigma(\cdot)$ is the sigmoid function. The slide-level representation is the attention-weighted sum $\mathbf{s} = \sum_{i=1}^{N} a_i\,\mathbf{z}_i$, passed to a binary classification head $\hat{y} = \sigma(\mathbf{w}_c^\top \mathbf{s} + b_c)$.

\noindent\textbf{Training.} Models were trained with focal loss ($\gamma{=}2.0$)~\cite{ref_focal}, AdamW (lr$=10^{-4}$, weight decay $10^{-3}$), cosine annealing over 100 epochs, and early stopping (patience 20). Evaluation used 5-fold patient-stratified cross-validation with out-of-fold (OOF) predictions.

\noindent\textbf{Multi-task abMIL.} Shared attention across all nine targets with task-specific classification heads. Architecture and training matched single-task except for fold stratification by TP53 status.
%-------------------------------------------------------------------------
\subsection{Spatial Transcriptomic Attention Validation}
\noindent\textbf{Data and alignment.} Visium spatial transcriptomics data from 18 GBM samples (${\sim}69{,}000$ spots) served as an independent validation set~\cite{ref_ravi}. We used the HEST-Library~\cite{ref_hest} to align H\&E images with Visium spot coordinates and extract $224{\times}224$ patches ($0.5\;\mu\text{m/pixel}$) centered at each spot, yielding one patch per spot. Each frozen encoder $f_\theta$ was then applied to these co-registered patches, producing spot-level embeddings in the same feature space as the CPTAC training patches.

\noindent\textbf{Attention mapping.} The set of spot-centered embeddings for each Visium sample was treated as a bag and passed through the trained abMIL checkpoints from CPTAC. For each encoder--task combination, all five fold checkpoints were applied and attention weights averaged across folds for robust estimates. Because HEST extracts one patch per spot, attention weights map directly to Visium spot coordinates via coordinate alignment, yielding a spot-level attention score $a_s$ for each spot $s$.

\noindent\textbf{Biological signatures.} We defined $K{=}87$ signatures spanning five hierarchical scales: 13 Hallmark pathways (MSigDB)~\cite{ref_msigdb}, 4 GBM subtype programs~\cite{ref_verhaak}, 14 GBM transcriptional programs~\cite{ref_greenwald}, 12 general cell-type gene sets, and 44 individual genes. For each signature, z-scores were computed on log-normalized expression per spot. The hallmark pathways were selected based on the most applicable to cancer research.

\noindent\textbf{Enrichment analysis.} For each encoder--task--sample triplet, we stratified spots into high-attention (top 10\%) and low-attention (bottom 10\%) groups and quantified enrichment via Cohen's d for a particular signature. Statistical significance was assessed via Welch's $t$-test with Benjamini--Hochberg FDR correction across all comparisons. Variance decomposition via Kruskal--Wallis $\eta^2$ quantified contributions of sample identity, encoder, and task. Spatial autocorrelation was assessed via Moran's $I$ on rank-transformed and raw attention.
\section{Results and Discussion}
\begin{table}[t]
\centering
\caption{Patient-level AUROC for single-task (ST) and multi-task (MT) models. 
Top: CPTAC 5-fold cross-validation. Bottom: TCGA external validation. 
\textbf{Bold}: best per column. \underline{Underline}: best mean.}
\label{tab:performance}
\setlength{\tabcolsep}{3pt}
\footnotesize
\begin{tabular}{@{}lccccccc@{}}
\toprule
 & & EGFR & TP53 & IDH1 & CDKN2A & PDGFRA & \\
Encoder & Task & Amp. & Mut. & Mut. & Del. & Amp. & Mean \\
\midrule
\multicolumn{8}{l}{\textbf{CPTAC Internal Validation (5-Fold CV)}} \\
\midrule
\multirow{2}{*}{ResNet50}    & ST & .673 & .601 & .478 & .557 & .514 & .565 \\
                              & MT & .702 & .580 & .455 & .652 & .602 & .598 \\
\multirow{2}{*}{CONCH v1.5}  & ST & .681 & \textbf{.734} & \textbf{.729} & .638 & .491 & .655 \\
                              & MT & .703 & \textbf{.742} & \textbf{.733} & .665 & .618 & \underline{\textbf{.692}} \\
\multirow{2}{*}{UNI v2}      & ST & .696 & .667 & .717 & .643 & .431 & .631 \\
                              & MT & .690 & .679 & .586 & .676 & .703 & .667 \\
\multirow{2}{*}{Virchow2}    & ST & .722 & .646 & .648 & .647 & .544 & .642 \\
                              & MT & .669 & .661 & .554 & .716 & .565 & .633 \\
\multirow{2}{*}{GigaPath}    & ST & .722 & .663 & .683 & .637 & .534 & .648 \\
                              & MT & .709 & .717 & .530 & \textbf{.718} & .672 & .669 \\
\multirow{2}{*}{H-Optimus-1} & ST & \textbf{.761} & .690 & .726 & \textbf{.674} & \textbf{.548} & \underline{\textbf{.680}} \\
                              & MT & \textbf{.730} & .659 & .646 & .667 & \textbf{.713} & .683 \\
\midrule
\multicolumn{8}{l}{\textbf{TCGA External Validation}} \\
\midrule
\multirow{2}{*}{ResNet50}    & ST & .592 & .621 & .905 & .669 & .588 & .675 \\
                              & MT & .566 & .621 & .852 & .665 & .501 & .641 \\
\multirow{2}{*}{CONCH v1.5}  & ST & .556 & .682 & .946 & .661 & .584 & .686 \\
                              & MT & .566 & .695 & .889 & .708 & .656 & .703 \\
\multirow{2}{*}{UNI v2}      & ST & .678 & \textbf{.684} & .984 & \textbf{.734} & \textbf{.713} & \underline{\textbf{.759}} \\
                              & MT & .659 & \textbf{.698} & .986 & .715 & .709 & .754 \\
\multirow{2}{*}{Virchow2}    & ST & .649 & .628 & .913 & .681 & .646 & .703 \\
                              & MT & .642 & .674 & .938 & .683 & .570 & .701 \\
\multirow{2}{*}{GigaPath}    & ST & .664 & .655 & \textbf{.992} & .671 & .608 & .718 \\
                              & MT & .646 & .619 & .975 & .714 & .621 & .715 \\
\multirow{2}{*}{H-Optimus-1} & ST & \textbf{.686} & .665 & .975 & .695 & .680 & .740 \\
                              & MT & \textbf{.714} & .658 & \textbf{.990} & \textbf{.731} & \textbf{.710} & \underline{\textbf{.761}} \\
\bottomrule
\end{tabular}
\end{table}

\subsection{Benchmarking: No Universal Best Encoder}
\noindent\textbf{Task-specific and training-specific benefits.} Single-task abMIL results on CPTAC data revealed task-specific encoder advantages (Table 1). H-Optimus-1 achieved the highest AUROC for EGFR amplification (0.761), CONCH v1.5 for TP53 (0.734) and IDH1 (0.730), while performance for PDGFRA amplification was near-chance across all encoders (best: 0.548). Comparing training paradigms, single-task outperformed multi-task for IDH1 (+9.0\% mean AUROC across encoders) and EGFR (+1.5\%), while multi-task provided beneficial regularization for PDGFRA (+14.4\%).

\noindent\textbf{External Validation Inverts Performance Rankings.}
TCGA external validation fundamentally changed the encoder landscape (Table 1). UNI v2 ranked 5th of 6 on CPTAC (mean AUROC 0.631), but rose to 1st on TCGA (0.759). A mean generalization gap of +0.093 excluding IDH1 which demonstrates superior feature transferability despite modest internal performance. H-Optimus-1 was the best balanced performer (\#4 CPTAC, mean 0.680; \#2 TCGA, mean 0.740), showing robust generalization (+0.014 gap excl. IDH1) without the extreme internal-external discordance of other encoders. GigaPath performed similarly regardless of training scheme and testing cohort with regards to encoder rankings. \noindent\textit{This has immediate practical implications: single-cohort benchmarks, even with proper cross-validation, can produce misleading encoder recommendations. As such we recommend evaluating on external cohorts and favoring encoders with demonstrated generalization.}
\begin{figure}[!t]
\centering
\includegraphics[width=\textwidth]{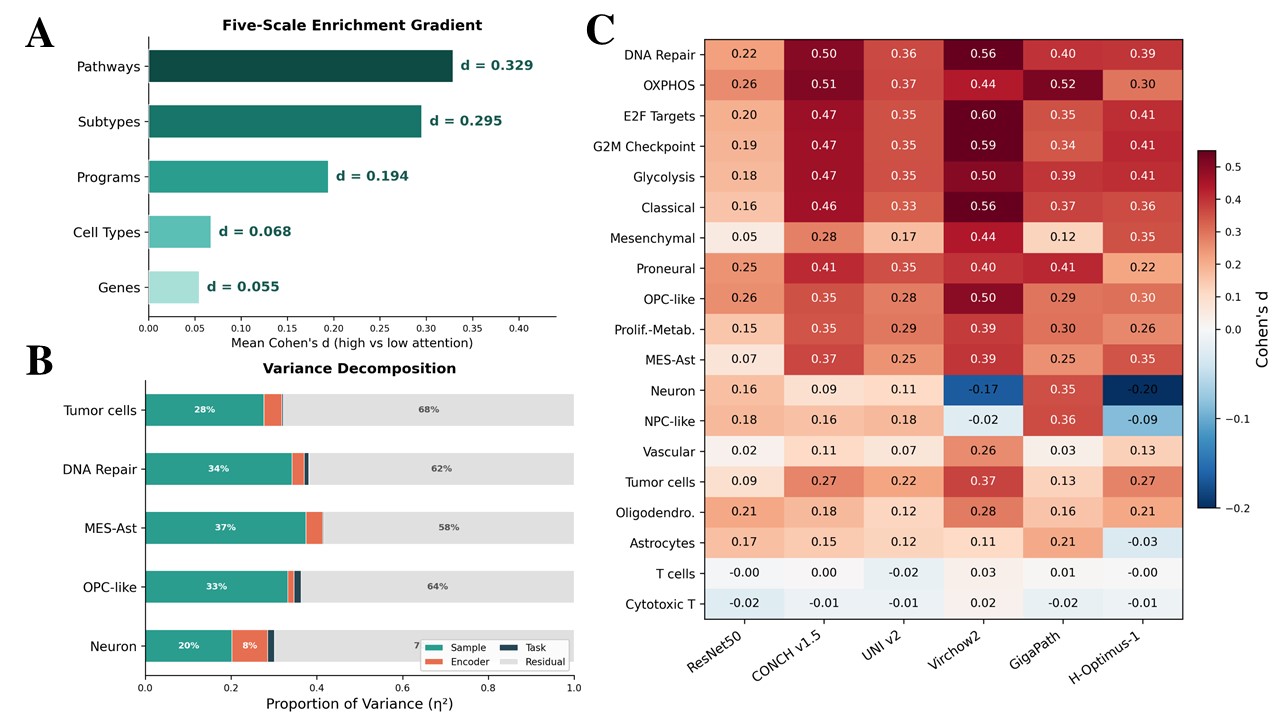}
\caption{\textbf{(A)}~Five-scale enrichment gradient from pathways to individual genes. \textbf{(B)}~Variance decomposition: sample heterogeneity dominates over encoder and task effects. \textbf{(C)}~Encoder-specific attention profiles across biological signatures.}
\end{figure}
\subsection{Orthogonal Validation via Spatial Transcriptomics Reveals Attention is Rooted in Biology}
\noindent\textbf{Attention Maps Capture Multi-Gene Biology.}
The spatial transcriptomic validation revealed that attention maps co-localize with transcriptional programs across all biological scales, but with a clear enrichment hierarchy (Fig 2A). Hallmark pathways showed the strongest enrichment (mean Cohen's d=0.329; 78\% FDR-significant), followed by GBM molecular subtypes (d=0.295; 79\%), transcriptional programs (d=0.194; 72\%), cell type signatures (d=0.068; 37\%), and individual genes (d=0.055; 25\%). This five-fold gradient from pathway to gene level supports that attention is more strongly associated with emergent multi-gene transcriptional states, and that  tissue-level biology is shaping histological morphology, rather than individual molecular markers. \noindent\textit{Attention maps should be evaluated against pathway-level or cell-state signatures as studies correlating attention with single biomarkers may underestimate the biological information encoded in attention patterns.}

\noindent\textbf{Metabolic-proliferative and GBM Transcriptional Program Selectivity.}
The most enriched pathways formed a coherent core: DNA repair (d=0.408), oxidative phosphorylation (d=0.398), E2F targets (d=0.395), G2M checkpoint (d=0.392), and glycolysis (d=0.383).  Among 14 GBM-specific programs, OPC-like cells showed one of the strongest enrichments (d=0.330), followed by proliferative-metabolic (d=0.289) and mesenchymal-astrocytic (MES-Ast; d=0.279). General cell type markers confirmed selective enrichment for tumor cells (d=0.224), oligodendrocytes (d=0.194), and astrocytes (d=0.120), with near-complete exclusion of adaptive immune populations: T cells (d=0.002), and cytotoxic T cells (d=-0.009) (Fig 2C). Models trained to predict alterations attend to the tumor compartment where these alterations originate, serving as an important internal consistency check. The pathways and GBM specific transcriptional programs converge on regions of active tumor cell division and metabolic reprogramming demonstrating the interpretation is grounded in biology~\cite{ref_ravi}. 

\noindent\textbf{Spatial coherence does not imply biological coherence.}
 Moran's I analysis of attention spatial autocorrelation revealed a striking dissociation between spatial and biological coherence (Fig 3). ResNet50 produced the most spatially contiguous attention maps (Moran's I = 0.595), 1.3-fold higher than the foundation model average (0.464). Yet ResNet50 showed the weakest biological enrichment: pathway-level Cohen's d of 0.161 versus a foundation model average of 0.363 (2.3-fold lower), and program-level d of 0.118 versus 0.210 (1.8-fold lower). Conversely, CONCH v1.5 exhibited the lowest spatial autocorrelation (Moran's I = 0.363) but among the highest pathway enrichment (d=0.403). This dissociation indicates that ResNet50 attention highlights large contiguous regions likely capturing gross morphological tumor-versus-normal architecture while foundation models attend to more spatially dispersed but biologically specific features that correspond to defined transcriptional programs (Fig 3A).
\begin{figure}[t]
\centering
\includegraphics[width=\textwidth]{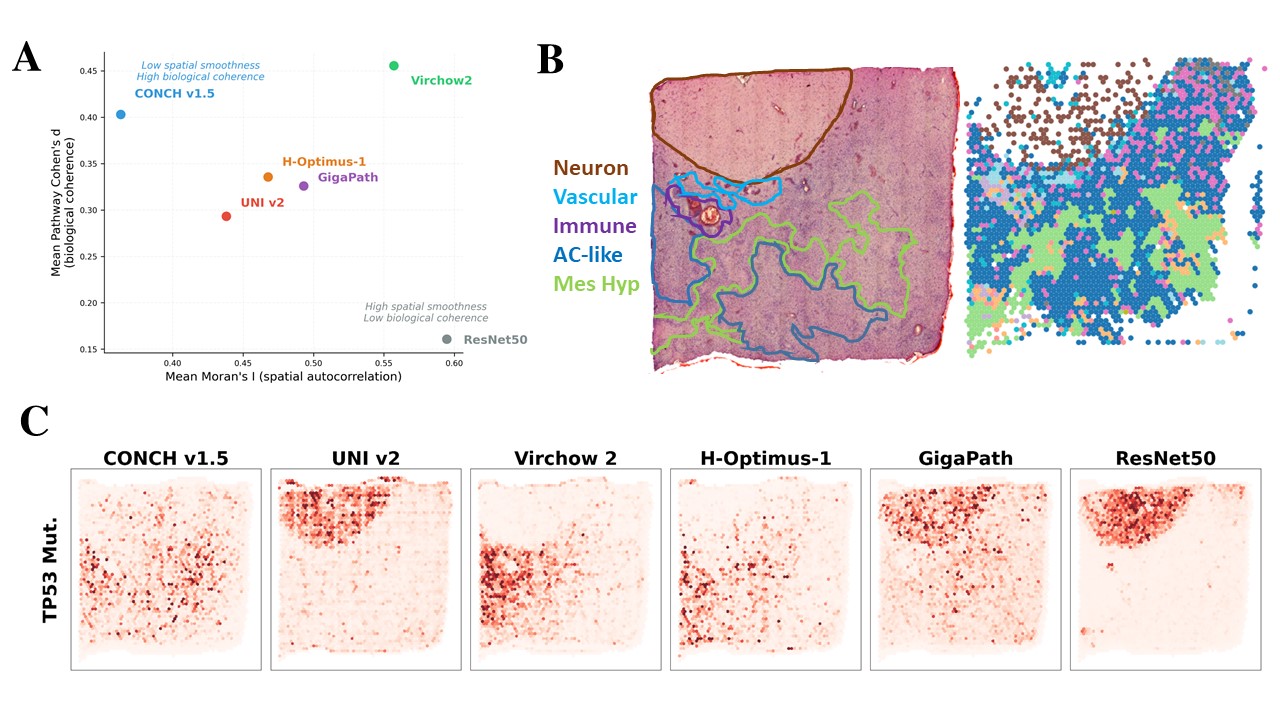}
\caption{\textbf{(A)}~Spatial autocorrelation (Moran's I) vs.\ biological enrichment reveals dissociation: ResNet50 is spatially smooth but biologically weak. \textbf{(B)}~Visium sample with transcriptomic compartment annotations. \textbf{(C)}~Encoder attention maps on a representative sample and task.}
\end{figure}
\subsection{Encoder-Specific Biological Profiles}
Despite contributing only 0.5-1.6\% of total variance globally, encoder choice produced biologically meaningful differences for specific programs. The Neuron signature demonstrated the most extreme encoder dependence (eta-squared=8\%). GigaPath exhibited uniquely strong neuronal affinity (d=0.355), with DE genes confirming large-effect synaptic marker enrichment (SNAP25 d=0.896). H-Optimus-1 showed essentially opposite attention on the Neuron program (d=-0.200), instead prioritizing glial and mesenchymal features (SPP1 d=0.748, BCAN d=1.222). Virchow2 showed a myelin profile (PLP1 d=0.925), while CONCH v1.5 was uniquely enriched for immune-related genes (HLA-B d=0.965). Sample-level heterogeneity dominated variance across all analyses, reflecting the biological reality of inter-tumor heterogeneity in GBM. The discovery that GigaPath attends to neuronal compartments (Neuron d=0.355) while H-Optimus-1 prioritizes glial features (Neuron, d=-0.200) suggests that spatial transcriptomic profiling of attention could serve as a principled approach for selecting encoder combinations and fusions. 
\section{Conclusion}
We propose a hypothesis-free attention evaluation framework and test it in a rigorous comprehensive benchmarking study with three key findings: (1) spatially coherent attention maps do not imply biological coherence and thus is a poor proxy for biological interpretability; (2) attention maps from foundation models capture biologically coherent multi-gene transcriptional programs in GBM; and (3) different foundation model encoders attend to distinct biological compartments, providing mechanistic insight into representational diversity. One of our limitations is that attention-transcriptome associations could reflect causal links to predicted alterations or correlations through shared morphological features. Our spatial transcriptomics evaluation framework offers a principled, quantitative approach for assessing what foundation models learn from histopathology that can be applied in any organ, moving the field beyond qualitative saliency map review.

\begin{credits}
\subsubsection{\ackname} 
This work was supported by the CIHR Vanier Scholarship (D.S) 

\subsubsection{\discintname}
The authors have no competing interests to declare that are relevant to the content of this article.
\end{credits}

\end{document}